\documentclass[conference]{IEEEtran}
\IEEEoverridecommandlockouts 
\usepackage{cite}
\usepackage{amsmath,amssymb,amsfonts}
\usepackage{graphicx}
\usepackage{textcomp}
\usepackage{xcolor}
\usepackage{subfig} 
\usepackage{algorithm} 
\usepackage{algpseudocode} 
\usepackage{algorithmicx}
\usepackage{booktabs}
\usepackage{pifont}
\usepackage{amssymb}
\usepackage{graphicx}
\usepackage{tabularx}
\usepackage{wasysym}
\usepackage{makecell}
\usepackage{amsmath}   
\usepackage{xcolor}  
\usepackage{svg}
\usepackage{colortbl} 
\usepackage{caption}
\usepackage{marvosym}

\def\BibTeX{{\rm B\kern-.05em{\sc i\kern-.025em b}\kern-.08em
    T\kern-.1667em\lower.7ex\hbox{E}\kern-.125emX}}

\definecolor{lighterpurple}{rgb}{0.92,0.92,0.98}

\begin{document}	
	\title{ElimPCL: Eliminating Noise Accumulation \\ with Progressive Curriculum Labeling for Source-Free Domain Adaptation}

	\author{
		\IEEEauthorblockN{
			Jie Cheng\textsuperscript{*},
			Hao Zheng\textsuperscript{*},
			Meiguang Zheng\textsuperscript{\Letter}, 
            Lei Wang, 
            Hao Wu, 
			Jian Zhang}
		\IEEEauthorblockA{School of Computer Science and Engineering, Central South University, Changsha, China}
		\IEEEauthorblockA{\{jiecheng, zhenghao, zhengmeiguang, wanglei, wuhaocs, jianzhang\}@csu.edu.cn}}
        
	\maketitle
	
	\begin{abstract}
		Source-Free Domain Adaptation (SFDA) aims to train a target model without source data, and the key is to generate pseudo-labels using a pre-trained source model.
		However, we observe that the source model often produces highly uncertain pseudo-labels for hard samples, particularly those heavily affected by domain shifts, leading to these noisy pseudo-labels being introduced even before adaptation and further reinforced through parameter updates. 
		Additionally, they continuously influence neighbor samples through propagation in the feature space.
		To eliminate the issue of noise accumulation, we propose a novel Progressive Curriculum Labeling (ElimPCL) method, which iteratively filters trustworthy pseudo-labeled samples based on prototype consistency to exclude high-noise samples from training. 
		Furthermore, a Dual MixUP technique is designed in the feature space to enhance the separability of hard samples, thereby mitigating the interference of noisy samples on their neighbors.
		Extensive experiments validate the effectiveness of ElimPCL, achieving up to a 3.4\% improvement on challenging tasks compared to state-of-the-art methods.
	\end{abstract}

        \begin{IEEEkeywords}
		Source-Free Domain Adaptation, Noise Accumulation, Curriculum Labeling
        \end{IEEEkeywords}

        \let\thefootnote\relax
        \footnotetext{\textsuperscript{*}Co-first authors. \textsuperscript{\Letter}Meiguang Zheng is the corresponding author.}
        \footnotetext{This work is supported by the National Natural Science Foundation of China (No. 62172442 and No. 62472446), China Scholarship Council, and High Performance Computing Center of Central South University.}

    \section{Introduction}
	Unsupervised domain adaptation (UDA)\cite{meng2022unsupervised, zhu2023srouda} focuses on transferring knowledge learned from a large labeled source domain to an unlabeled target domain, which helps reduce the cost of data collection and labeling. 
	Standard UDA methods typically require access to both the source and target data to address performance degradation caused by domain shifts. 
	However, this may not be feasible in many applications, particularly when data privacy or transmission bandwidth are of critical importance. 
	To overcome the reliance of standard UDA on source data, source-free domain adaptation (SFDA)\cite{liang2020we, liang2022domain} has been proposed in recent years.
	SFDA aims to adapt a pre-trained source model to an unlabeled target domain without needing access to the source data.
	
    \begin{figure}[ht]
		\centering
		\includegraphics[width=0.48\textwidth, height=0.25\textwidth]{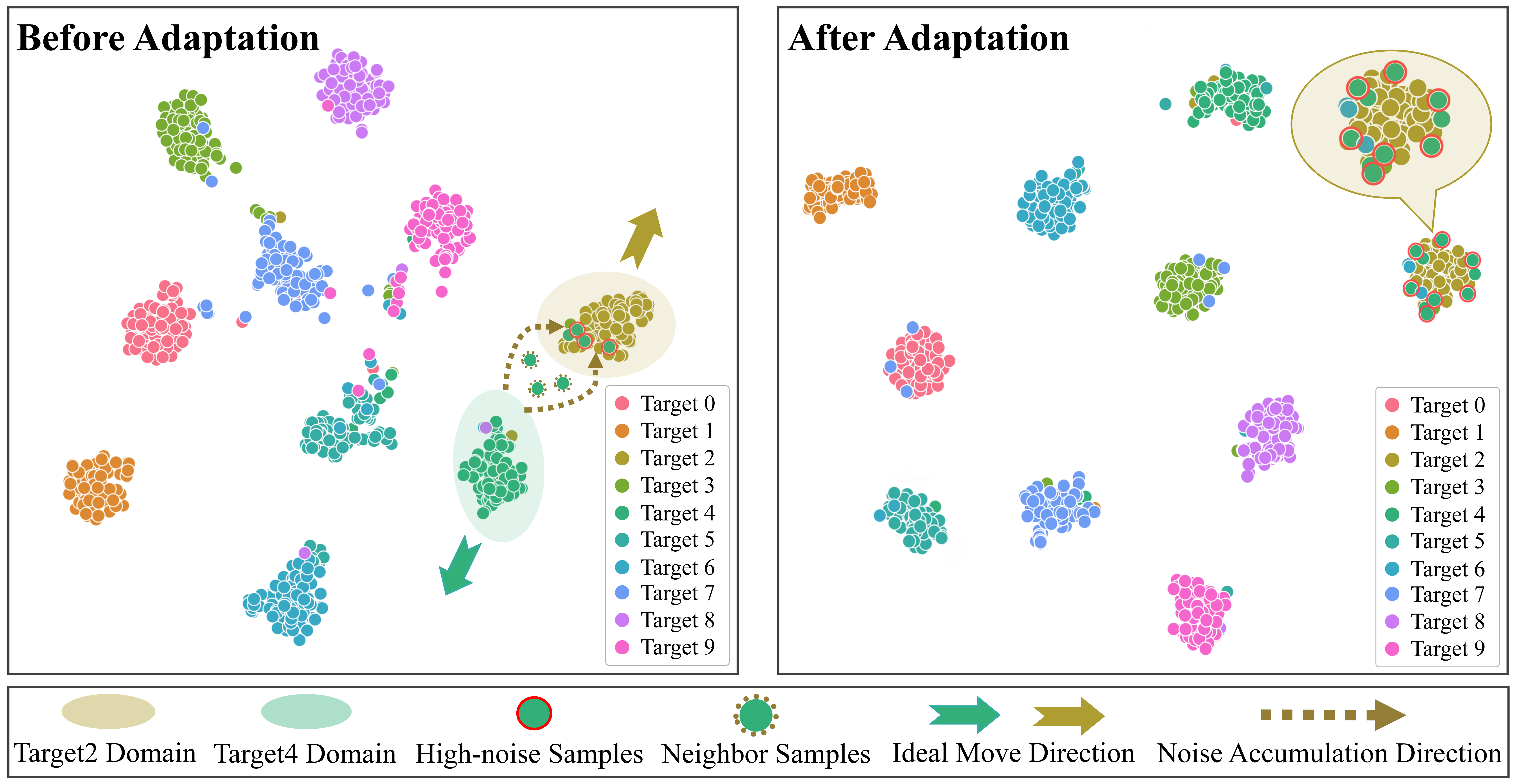}
		\caption{An illustration of the noise accumulation issue on \textbf{\textit{Office-Caltech}} Caltech$\rightarrow$Amazon.
		A few high-noise samples have been extremely misaligned before adaptation due to heavy domain shifts.
		Additionally, they induce neighbor samples to be misaligned as well during domain adaptation.
		This leads to accumulating a large number of noisy samples.}
            \vspace{-0.5cm}
		\label{problem}
	\end{figure}

	Recent SFDA methods mainly leverage the self-supervised learning paradigm to further fine-tune the pre-trained source model. 
	However, these methods generally focus on generating high-quality pseudo-labels for unlabeled target samples using Noisy Label Learning (NLL) techniques during domain adaptation. 
	In practice, the source model often generates highly uncertain pseudo-labels for hard samples before domain adaptation, which cannot be fully denoised due to significant domain shifts. 
	Since the source model heavily relies on clean labels in the early training stages, the negative transfer effects caused by these high-noise pseudo-labels can misguide model optimization and accumulate over the course of training. 
	
        To further illustrate the accumulation problem, we visualize the feature space before and after domain adaptation using conventional NLL techniques. 
	As shown in Fig.\ref{problem}, we observe that some high-noise samples are already introduced before adaptation, particularly those that are heavily affected by domain shifts. 
	Even more concerning is that their pseudo-labels are reinforced due to the biased optimization direction, making the source model more confident in its incorrect predictions. 
	Moreover, neighbor samples in the feature space are also influenced by these high-noise samples through noise propagation.
	For instance, some Target4 samples in Fig.\ref{problem} were already misaligned to the Target2 domain before adaptation due to their high confidence pseudo-labels.
	These high-noise samples not only are constantly reinforced through parameter updates, 
	but also cause neighbor samples, which could have been correctly classified, to become misaligned as well. 
	This reinforcement and propagation of noise create a vicious cycle, continuously amplifying the noise accumulation problem and ultimately leading to model collapse.

	To eliminate noise accumulation, we propose a novel Progressive Curriculum Labeling (ElimPCL) approach for SFDA, 
	where the pre-trained source model begins with easy samples and gradually progresses to harder ones.
	First, a curriculum is carefully designed based on prototype consistency to guide a student model learning from trustworthy pseudo-labeled samples iteratively.
	In this way, high-noise samples are excluded from training before adaptation, avoiding noisy pseudo-labels being reinforced during subsequent training.
	Second, a new Dual MixUP technique, consisting of Intra-MixUP and Inter-MixUP, is designed in the feature space to mitigate the propagation of mislabeled information during adaptation.
	Dual MixUP gradually promotes the hard samples to become separable under the guidance of pairwise structural information in the target domain.
	Finally, the source model is further fine-tuned by fusing the student model parameters using an adaptive smooth parameter movement method.
	
	Our main contributions are summarized as follows:
	\begin{itemize}
        \item We identify the noise accumulation phenomenon before adaptation and propose an effective ElimPCL method that leverages prototype consistency to progressively guide the model from easy to harder samples, ensuring that high-noise samples are excluded prior to adaptation.
        \item We design a Dual MixUP data augmentation technique in the feature space to facilitate the source model extracting more discriminative features for hard samples during adaptation, thus preventing the noisy pseudo-labels being spread to neighbor samples.
		\item We validate the effectiveness of our proposed method on four benchmark datasets. Especially for the transfer tasks with severe domain shifts, ElimPCL exhibits more powerful competitive advantage.
	\end{itemize}

	\section{Related Work}
	\noindent \textbf{Unsupervised Domain Adaptation (UDA).} 
	Standard UDA methods aim to adapt a model trained on source domain to an unlabeled target domain, allowing the knowledge from the source domain to be transferred adaptively to the target domain without additional annotation.
	Recent UDA methods are mainly divided into two main directions: discrepancy-based\cite{zhu2020deep, qiu2023rkhs, SunXL023} and adversarial learning\cite{ganin2015unsupervised, li2021bi, saito2018maximum}.
	Discrepancy-based methods focus on aligning the labeled source data and unlabeled target data within a shared representation space.
	Adversarial learning minimizes the domain gap by adding a gradient reversal layer.
	Overall, these standard UDA methods all require sufficiently labeled source domain data to realize domain alignment and classification.

	\noindent \textbf{Source-Free Domain Adaptation (SFDA).} 
	SFDA aims to adapt a pre-trained source model to an unlabeled target domain without accessing the source data.
	Data-driven methods concentrate on reconstructing the source domain, then aligning the target domain to the virtual source domain using the standard UDA methods \cite{tian2021vdm, YehYYH21, ding2023proxymix, xia2021adaptive}. 
	However, using generative models is not only computationally expensive but also raises significant challenges for domain generalization.
	Model-driven methods usually fine-tune the source model using self-supervised learning techniques\cite{yang2021exploiting, pei2023uncertainty, tang2024source, DingXT0WT22, chen2022self, zheng2023multifeature,zheng2022metaboost}.
	These methods typically focus on refining pseudo-labels generated by the pre-trained source model during domain adaptation, overlooking the negative transfer effects caused by high-noise samples before adaptation.

	\section{Proposed Method}
    We consider a typical source-free domain adaptation task for image classification, where the source domain $D_s$ consists of pairs of images and ground-truth labels $\{x_s, y_s\}$, with $x_s \in \mathcal{X}_s$ and $y_s \in \mathcal{Y}_s$.
	Let $D_t$ be the target domain only including images $\{x_t\}$, where $x_t \in \mathcal X_t$.
	In the SFDA setting, we are only given a pre-trained source model $f_{\theta_s}$, without having access to the original source domain $D_s$.
    The source model $f_{\theta_s}: \mathcal X_s \rightarrow \mathcal Y_s$, with $\theta_s = \{g_s, h_s\}$, consists of a typical architecture composed by a feature extractor $g_s:\mathcal X_s \rightarrow \mathbb{R}^d$ and a classifier $h_s:\mathbb{R}^d \rightarrow \mathbb{R}^K$, where $d$ is the dimension of the input features and $K$ is the number of classes.
	
	At the beginning of the adaptation process, the pre-trained source model $f_{\theta_s}$ is used to generate pseudo-labels $\hat{\mathcal Y}_t$ for the unlabeled target data $\mathcal X_t$: 
	\begin{equation}
		\begin{aligned}
			\hat{\mathcal Y}_t = \arg\max f_{\theta_s}(\mathcal X_t)
		\end{aligned}
	\end{equation}

	Due to domain shifts, the source model consistently makes incorrect predictions, which can be interpreted as noise in the pseudo-labels. 
	Therefore, the goal of the adaptation phase is to gradually refine these noisy pseudo-labels, thereby adapting the source model to the target domain:
	\begin{equation}
		\begin{aligned}
			\min_{f_{\theta_s}} \mathcal L(f_{\theta_s} \mid \mathcal X_t, \hat{\mathcal Y}_t)
		\end{aligned}
	\end{equation}

\begin{figure*}[ht]
		\begin{center}
			\includegraphics[width=0.95\textwidth, height=0.3\textwidth]{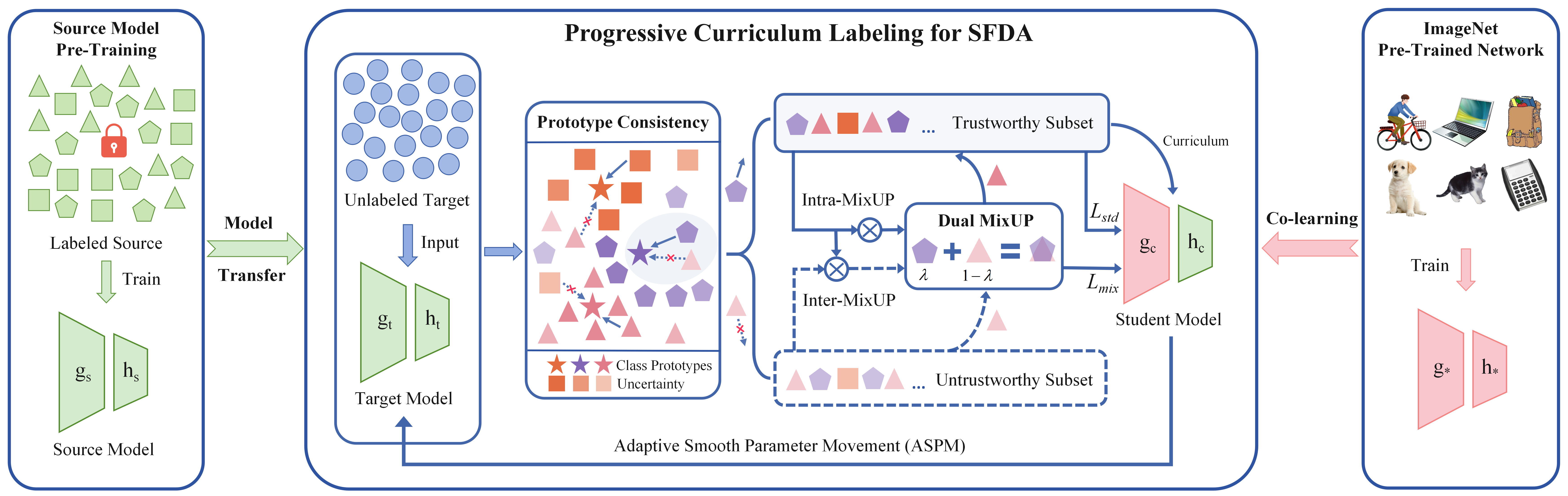}
		\end{center}
            \vspace{-0.3cm}
		\caption{Overview of ElimPCL. 
		The pseudo-labels generated by the source model are first fed into the prototype consistency module to divide the target domain into a trustworthy and untrustworthy subset.
		The trustworthy subset is used as the curriculum to guide a student model training, excluding interference from high-noise samples.
		Then, these two sets of samples are mixed with features and curriculum-labels simultaneously using Dual MixUP to facilitate feature learning for hard samples.
		Finally, the source model is fine-tuned via fusing the student model parameters by co-learning with ImageNet pre-trained network.}
            \vspace{-0.5cm}
		\label{architecture}
	\end{figure*}

	However, the pre-trained source model often generates extremely unreliable pseudo-labels for hard samples before adaptation.
	If these high-noise samples participate in domain adaptation too early, it will mislead the optimization direction of source model.
	As shown in Fig.\ref{architecture}, we propose a novel ElimPCL method for SFDA to alleviate noise accumulation caused by noise reinforcement and propagation effects.
	
	\subsection{Curriculum Design Based on Prototype Consistency}
	To exclude high-noise samples before adaptation, we design a curriculum based on prototype consistency to guide a student model learning from trustworthy pseudo-labeled samples, which can be divided into three steps: Prototype Generation, Pseudo-label Refinement, and Consistency Filtering.
	
	\noindent \textbf{Prototype Generation}.
	Class prototypes provide a robust and reliable representation of the distribution of different categories within the target domain.
	We calculate the centroid of each class, which can be used as class prototypes:
	\begin{equation}
		\begin{aligned}
			c_k = \frac{\sum_{x_t \in \mathcal X_t} \delta_k(f_{\theta_s}(x_t)) \cdot g_s(x_t)}{\sum_{x_t \in \mathcal X_t} \delta_k(f_{\theta_s}(x_t))}
		\end{aligned}
	\end{equation}

	\noindent where $c_k$ denotes the prototype of the $k$th class, $\delta_k$ denotes the $k$th element in the softmax operation.

	\noindent \textbf{Pseudo-label Refinement}. 
	Once the class prototypes are obtained, the pseudo-label of each sample can be refined by its nearest or most similar class prototype:
	\begin{equation}
		\begin{aligned}
			\tilde{y}_t = \arg\min_k D_{cos}(g_s({x}_t), c_k)
		\end{aligned}
	\end{equation}
	\noindent where $\tilde{y}_t$ denotes the refined pseudo-label of ${x}_t$, we take cosine similarity to measure the distance between $x_t$ and each class prototype $c_k$.

	\noindent \textbf{Consistency Filtering}.
        The high uncertainty of pseudo-label $\hat{y}_t$ means that it may contain high-noise.
	Therefore, we first leverage entropy to estimate the uncertainty of $\hat{y}_t$ as follows:
    	\begin{equation}
		\begin{aligned}
			\mathcal{H}(\hat {y}_t) &= -\sum_{k=1}^{K} \hat{y}_t^{k} \log \hat{y}_t^{k}\\
		\end{aligned}
	\end{equation}
        where the lower value of $\mathcal{H}(\hat {y}_t)$ indicates that $\hat{y}_t$ is more reliable.
        On the other hand, if the target pseudo-label $\hat{y}_t$ is consistent with the refined pseudo-label $\tilde{y}_t$, the noise can be effectively eliminated.
	Such samples are defined as easy samples, indicating minimal impact from domain shifts, while others are classified as hard samples.
	The target domain is then divided into a trustworthy subset $D_{tt}$ and an untrustworthy subset $D_{ut}$:
	\begin{equation}
		\begin{aligned}
			D_{tt} = \{({x}_t, &\hat {y}_t) \mid \mathcal{H}(\hat {y}_t) < \tau, \hat {y}_t = \tilde {y}_t\} \label{split_eq}
		\end{aligned}
	\end{equation}
	\noindent where $\tau$ is the uncertainty threshold. The remaining samples are divided into $D_{ut}$. The trustworthy subset $D_{tt}$ is used as the curriculum to train a student model $f_{\theta_c}$ as shown in Fig.\ref{architecture}.
	The cross-entropy loss function is calculated as follows: 
	\begin{equation}
		\begin{aligned}
			\mathcal L_{std}^{ce} = - \mathbb{E}_{(x_t, \hat {y}_t) \in D_{tt}} \sum_{k=1}^{K} q_k \log \delta_k(f_{\theta_c}(x_t))
		\end{aligned}
	\end{equation}
	\noindent where $\{q_k\}_{k=1}^K$ is the one-hot vector of  $\hat {y}_t$.
    $L_{std}^{ce}$ avoids high-noise samples in $D_{ut}$ participating in adaptation too early.
	
	To further reduce the uncertainty of student model predictions, we apply the following regularization term to guide the target outputs similar to one-hot coding:
	\begin{equation}
		\begin{aligned}
			\mathcal L_{std}^{ent} = - \mathbb{E}_{(x_t, \hat {y}_t) \in D_{tt}} \sum_{k=1}^{K} \delta_k (f_{\theta_c}(x_t)) \log \delta_k (f_{\theta_c}(x_t))
		\end{aligned}
	\end{equation}
	
	The loss function employed for training the student model is as follows:
	\begin{equation}
		\begin{aligned}
			\mathcal L_{std} =\gamma\mathcal L_{std}^{ce} + \mathcal L_{std}^{ent} \label{student_eq}
		\end{aligned}
	\end{equation}
	\noindent where $\gamma$ is a non-tuned hyperparameter that regulates the balance between the cross-entropy loss function and the regularization term. $L_{std}$ can guide the student model learning from easy to harder samples.
	
	\subsection{Dual MixUP with Restricted Mixing Ratio}
	As shown in Fig.\ref{architecture}, a new data augmentation technique Dual MixUP in the feature space is proposed to enhance the separability of hard samples, which consists of Intra-MixUP and Inter-MixUP.
	By mixing features from the trustworthy subset $D_{tt}$ samples, the source model is accelerated to capture the discriminative features for the untrustworthy subset $D_{ut}$ samples during domain adaptation.
	
	Dual MixUP considers the pairwise structural information in the target domain.
	The new sample $\{\tilde{x},\tilde{y}\}$ generated by MixUP $(x_1,\hat{y}_1)$, $(x_2,\hat{y}_2)$ can be defined as:
	\begin{equation}
		\begin{cases}
			\begin{aligned}
				\tilde{x} &= \lambda x_1 + (1-\lambda) x_2 \\
				\tilde{y} &= \lambda \hat{y}_1 + (1-\lambda) \hat{y}_2
			\end{aligned}
		\end{cases}
	\end{equation}
	\noindent where $\lambda \sim Beta(\alpha, \alpha)$ is a mixing ratio which is randomly sampled from a $Beta$ distribution, $\alpha$ is a hyperparameter. 
	For the Intra-MixUP, as $D_{tt}$ contains more easy samples and the pseudo-labels are already very clean, we randomly mix within the subset of $D_{tt}$ samples.

	For the Inter-MixUP between $D_{tt}$ and $D_{ut}$, as $D_{ut}$ contains part of noisy hard samples, there is still a certain amount of hidden noisy labels.
	Therefore, the mixing ratio $\lambda$ needs to be limited to avoid confusing the classification boundaries of other classes. 
	The intuitive understanding is that if the label prediction quality of the source model is high, 
	it would be better that the value of $\lambda$ is closer to $0.5$, which facilitates the generation of more diverse intermediate feature representations.
	Conversely, if the overall pseudo-labels quality is poor, $\lambda$ should be closer to $1.0$, mixing more $D_{tt}$ samples, to avoid injecting more noise. 
	However, the accuracy of the source model cannot be measured based on ground-truth labels directly. 
	We argue that the ratio between $|D_{tt}|$ and $|D_{ut}|$ to some extent reflects the performance of the source model:
	\begin{equation}
		\begin{aligned}
			r = \frac{|D_{tt}|}{|D_{tt}|+|D_{ut}|}
		\end{aligned}
	\end{equation}
	where the value of $r$ is between $[0,1]$, the higher value and the better performance of the source model.
	Moreover, the restricted distribution parameter $\hat{\alpha}$ and mixing ratio $\hat{\lambda}$ can be expressed as:
	\begin{equation}
		\begin{cases}
			\begin{aligned}
				\hat{\alpha} &= \alpha r^2 \\
				\hat{\lambda} \sim &Beta(\hat{\alpha}, \hat{\alpha}) \label{lambda_eq}
			\end{aligned}
		\end{cases}
	\end{equation}

	Then the student model can be further refined by the new Dual MixUP samples $\{\tilde{x},\tilde{y}\}$:
	\begin{equation}
		\begin{aligned}
			\mathcal L_{mix}(f_{\theta_c} \mid \tilde{x}, \tilde{y}) = -\mathbb{E}_{x_1, x_2 \in \mathcal X_t} \sum_{k=1}^{K} \tilde{q}_k \log \delta_k(f_{\theta_c}(\tilde{x})) \label{mix_eq}
		\end{aligned}
	\end{equation}
	\noindent where $\{\tilde{q}_k\}_{k=1}^K$ is the one-hot vector of mixing label $\tilde{y}$.
    $L_{mix}$ promotes the discriminability of hard samples in the feature space, thus effectively mitigating noise propagation.

	\subsection{Fine-tuning the Source Model under Curriculum}
	Typically, the pre-trained ImageNet weights are usually applied to initialize the source model and subsequently discarded.
	Actually, the larger and more diverse pre-training dataset is not source-biased and may better capture the target input distribution.
	As depicted in Fig.\ref{architecture}, a co-learning strategy is proposed to adapt the student model by integrating the generalized feature extraction capability of the ImageNet pre-trained network.
	Since the classifier of the pre-trained network is not suitable for the current task, the student classifier is still initialized by the source classifier, i.e. $g_c=g_*, h_c=h_s$.
	Then, the student model is updated by minimizing the total loss:
	\begin{equation}
		\begin{aligned}
			\mathcal L_{tot} =\mathcal L_{std} + \mu \mathcal L_{mix}
		\end{aligned}
	\end{equation}
	where $\mu$ is a loss coefficient that controls the pace of curriculum training. 
	Since DNNs tend to first memorize correctly labeled data before incorrectly labeled samples, $\theta_c$ is reinitialized after each epoch to further mitigate noise accumulation.

	Subsequently, the source model is fine-tuned by fusing the student model parameters with a smoothing ratio $\beta$.
	Considering that the pseudo-labels of hard samples are gradually refined as the adaptation process goes deeper, the trustworthy subset $D_{tt}$ expands. 
	The performance of the student model gradually improves, eventually surpassing that of the source model.
	It would be better to let $\beta$ change dynamically according to the state of the student model.
	Therefore, the source model is refined using an improved Adaptive Smooth Parameter Movement (ASPM) method, the latest source model parameters $\theta_n$ after the $n$th epoch is as follows:
	\begin{equation}
		\begin{aligned}
			\theta_n \leftarrow \beta_n \theta_c + (1-\beta_n) \theta_{n-1} \label{source_eq}
		\end{aligned}
	\end{equation}
	\noindent where $N$ is the total target epochs, $\beta_n$ adaptively increases in $\Delta = \frac{\beta_{N}-\beta_{0}}{N}$ increments, i.e. $\beta_n = \beta_{n-1} + \Delta$.
	For clarity, we also summarize the overall training process of proposed ElimPCL method in Algorithm \ref{algo1}.
	
	\begin{algorithm}[t]
		\renewcommand{\algorithmicrequire}{\textbf{Input:}}
		\renewcommand{\algorithmicensure}{\textbf{Output:}}
		\caption{Overall training of ElimPCL.}
		\label{algo1}
		\small
		\begin{algorithmic}[1]
			\Require The source model $\theta_s$, the student model $\theta_c$, the ImageNet pre-trained network $\theta_*$, target data $\mathcal X_t$, target epochs $N$, sub epochs $K$, 
			smoothing ratio $\beta_{0}$ and $\beta_{N}$.
			\Ensure The target model $\theta_t = \theta_N$
	    
			\State Let $\theta_0 = \theta_c = \theta_s$, $\Delta = \frac{\beta_{N}-\beta_{0}}{N}$
			\For {$n=1\ to\ N$}
			  \State Re-initialize the student model: $\theta_c = \{g_*, h_s\}$
			  \State Split $\mathcal X_t$ into $D_{tt}$ and $D_{ut}$ based on prototype consistency
			  
			  \For {$k=1\ to\ K$}
			    \State Update the student model using $D_{tt}$ by Eq.\ref{student_eq}
			  \EndFor
                \State Intra-MixUP and Inter-MixUP by Eq.\ref{mix_eq}

			  \State $\beta_n = \beta_{n-1} + \Delta$
			  \State Fine-tune the source model: $\theta_n \leftarrow \beta_n \theta_c + (1-\beta_n) \theta_{n-1}$ 
			\EndFor
	 \end{algorithmic}
	\end{algorithm}

	\section{Experiments and Results}
	\subsection{Experimental Settings}
	\noindent \textbf{Datasets.} We conduct extensive experiments on \textbf{\textit{Office-31}}\cite{saenko2010adapting}, \textbf{\textit{Office-Caltech}}\cite{gong2012geodesic}, 
    \textbf{\textit{VisDA-C}}\cite{peng2017visda}, and \textbf{\textit{Digits}}\cite{liang2020we}.
	\textit{Office-31} contains three domains: Amazon(A), Webcam(W), and DSLR(D). 
	\textit{Office-Caltech} consists of four domains: Amazon(A), Webcam(W), DSLR(D), and Caltech256(C).
	\textit{VisDA-C} is a large-scale dataset from simulators to realistic environments, which consists of 12 common classes.
	\textit{Digits} contains three domains: SVHN(S), MNIST(M), and USPS(U). 

	\noindent \textbf{Implementation details.} Following SHOT\cite{liang2020we}, we use LeNet backbone for \textit{Digits}, ResNet50 backbone for \textit{Office-31} and \textit{Office-Caltech}, and ResNet101 backbone for \textit{VisDA-C}.
    For \textit{Digits}, we use the Adam optimizer to train all networks with a fixed weight decay $1e^{-4}$, a learning rate $\eta_0=2e^{-4}$ and a batch size 128.
	For \textit{Office-31} and \textit{Office-Caltech}, we adopt SGD optimizer with a momentum 0.9, a weight decay $1e^{-4}$, a learning rate $\eta_0=5e^{-3}$ and a batch size 64.
	For \textit{VisDA-C}, we also use SGD optimizer with a momentum 0.9, a weight decay $1e^{-4}$, a learning rate $\eta_0=3e^{-4}$ and a batch size 64.
	We set $\beta_{0}=0.3$, $\beta_{N} = 0.8$ in all experiments.
	For Intra-MixUP, $\alpha$ is set to 1.0, while for Inter-MixUP, $\alpha$ is initially set to 2.0.
	
	\begin{table*}[ht]
		\centering
		\caption{Classification accuracies(\%) on \textbf{\textit{VisDA-C}}. Bold font denotes the best results.}
		\label{VisDA-C}  
		\scriptsize
		\resizebox*{0.98\textwidth}{0.165\textheight}{
		 \begin{tabular}{cc|cccccccccccc>{\columncolor{lighterpurple}}c}
			\toprule	
			\textbf{Methods} & \textbf{Source} & \textbf{Plane} & \textbf{Bcycl} & \textbf{Bus} & \textbf{Car} & \textbf{Horse} & \textbf{Knife} & \textbf{Mcycl} & \textbf{Person} & \textbf{Plant} & \textbf{Sktbrd} & \textbf{Train} & \textbf{Truck} & \textbf{AVG.}   \\
			\noalign{\smallskip}\hline\noalign{\smallskip}
			DANN\cite{ganin2015unsupervised}      & \checked  & 81.9 & 77.7 & 82.8 & 44.3 & 81.2 & 29.5 & 65.1 & 28.6 & 51.9 & 54.6 & 82.8 & 7.8  & 57.4 \\
			MCD\cite{saito2018maximum}            & \checked  & 87.0 & 60.9 & 83.7 & 64.0 & 88.9 & 79.6 & 84.7 & 76.9 & 88.6 & 40.3 & 83.0 & 25.8 & 72.0 \\
			DSAN\cite{zhu2020deep}                & \checked  & 90.9 & 66.9 & 75.7 & 62.4 & 88.9 & 77.0 & 93.7 & 75.1 & 92.8 & 67.6 & 89.1 & 39.4 & 76.6 \\
			\noalign{\smallskip}\hline\noalign{\smallskip}
			SHOT\cite{liang2020we}                & $\times$   & 94.3 & 88.5 & 80.1 & 57.3 & 93.1 & 94.9 & 80.7 & 80.3 & 91.5 & 89.1 & 86.3 & 58.2 & 82.9 \\
			NRC\cite{yang2021exploiting}          & $\times$   & 96.8 & 91.3 & 82.4 & 62.4 & 96.2 & 95.9 & 86.1 & 80.6 & 94.8 & \textbf{94.1} & 90.4 & \textbf{59.7} & 85.9 \\
			A$^{2}$Net\cite{xia2021adaptive}      & $\times$   & 94.0 & 87.8 & 85.6 & 66.8 & 93.7 & 95.1 & 85.8 & 81.2 & 91.6 & 88.2 & 86.5 & 56.0 & 84.4 \\
			UTR\cite{pei2023uncertainty}          & $\times$   & \textbf{98.0} & \textbf{92.9} & 88.3 & 78.0 & \textbf{97.8} & \textbf{97.7} & 91.1 & \textbf{84.7} & 95.5 & 91.4 & 91.2 & 41.1 & 87.3 \\
			TPDS\cite{tang2024source}             & $\times$   & 97.6 & 91.5 & \textbf{89.7} & 83.4 & 97.5 & 96.3 & 92.2 & 82.4 & 96.0 & \textbf{94.1} & 90.9 & 40.4 & 87.7 \\
			\noalign{\smallskip}\hline\noalign{\smallskip}
			\textbf{ElimPCL(0urs)}                             & $\times$   & 95.2 & 82.0 & 83.8 & \textbf{85.6} & 96.3 & 96.4 & \textbf{94.5} & 81.9 & \textbf{98.3} & 91.2 & \textbf{95.3} & 54.1 & \textbf{87.9} \\
			\bottomrule
		 \end{tabular}
		}
        \vspace{-0.3cm}
	\end{table*}
	
	\subsection{Results and Comparison with SOTA}
	
	\begin{table}[ht]
		\caption{Classification accuracies(\%) on \textbf{\textit{Office-31}}.}
            \vspace{-0.2cm}
        \begin{center}
		\resizebox{0.48\textwidth}{0.067\textheight}{
		\begin{tabular}{cc|cccccc>{\columncolor{lighterpurple}}c}  
			\toprule	
			\textbf{Methods} & \textbf{Source\textbf} & \textbf{A$\rightarrow$D} & \textbf{A$\rightarrow$W} & \textbf{D$\rightarrow$A} & \textbf{D$\rightarrow$W} & \textbf{W$\rightarrow$A} & \textbf{W$\rightarrow$D} & \textbf{AVG.}  \\
			\noalign{\smallskip}\hline\noalign{\smallskip}
			DANN\cite{ganin2015unsupervised}      & \checked  & 79.7 & 82.0 & 68.2 & 96.9 & 67.4 & 99.1  & 82.2 \\
			DSAN\cite{zhu2020deep}                & \checked  & 90.2 & 93.6 & 73.5 & 98.3 & 74.8 & 100.0 & 88.4 \\
			SRoUDA\cite{zhu2023srouda}            & \checked  & 91.9 & 95.9 & 72.4 & 96.7 & 67.1 & 100.0 & 87.3 \\
			\noalign{\smallskip}\hline\noalign{\smallskip}
			SHOT\cite{liang2020we}                & $\times$   & 94.0 & 90.1 & 74.7 & 98.4 & 74.3 & 99.9  & 88.6 \\
			NRC\cite{yang2021exploiting}          & $\times$   & 96.0 & 90.8 & 75.3 & 99.0 & 75.0 & \textbf{100.0} & 89.4 \\
			A$^{2}$Net\cite{xia2021adaptive}      & $\times$   & 94.5 & 94.0 & 76.7 & \textbf{99.2} & 76.1 & \textbf{100.0} & 90.1 \\
			UTR\cite{pei2023uncertainty}          & $\times$   & 95.0 & 93.5 & 76.3 & 99.1 & \textbf{78.4} & 99.9  & 90.4 \\
			TPDS\cite{tang2024source}             & $\times$   & \textbf{97.1} & \textbf{94.5} & 75.7 & 98.7 & 75.5 & 99.8  & 90.2 \\
			\noalign{\smallskip}\hline\noalign{\smallskip}
			\textbf{ElimPCL(0urs)}                             & $\times$   & 94.9 & 91.7 & \textbf{80.1} & \textbf{99.2} & 77.3 & \textbf{100.0} &\textbf{90.5} \\
			\bottomrule
	\end{tabular} }
    \label{Office-31}  
    \end{center} 
    \vspace{-0.3cm}
	\end{table}

	\begin{table}[ht]
		\centering
		\caption{Classification accuracies(\%) on \textbf{\textit{Digits}}.}
		\label{Digits}  
		\scriptsize
		\resizebox{0.45\textwidth}{0.06\textheight}{
		\begin{tabular}{cc|ccc>{\columncolor{lighterpurple}}c}
			\toprule	
			\textbf{Methods} & \textbf{Source} & \textbf{S$\rightarrow$M} & \textbf{U$\rightarrow$M} & \textbf{M$\rightarrow$U} & \makebox[0.06\textwidth][c]{\textbf{AVG.}}   \\
			\noalign{\smallskip}\hline\noalign{\smallskip}
			DANN\cite{ganin2015unsupervised}  & \checked  & 73.8  & 85.1  & 73.0  & 77.3 \\
			MCD\cite{saito2018maximum}        & \checked  & 96.2  & 94.1  & 96.5  & 95.6 \\
			DSAN\cite{zhu2020deep}            & \checked  & 90.1  & 95.3  & 96.9  & 94.1 \\
			SRoUDA\cite{zhu2023srouda}        & \checked  & 88.7  & 98.5  & 95.0  & 94.1 \\
			SHOT\cite{liang2020we}            & $\times$  & 98.9  & 98.4  & 98.0   & 98.4 \\
			TPDS\cite{tang2024source}         & $\times$  & 98.9  & 97.8  & \textbf{98.4} & 98.4 \\
			\noalign{\smallskip}\hline\noalign{\smallskip}
			\textbf{ElimPCL(0urs)}                         & $\times$  & \textbf{99.1} & \textbf{98.7} & 98.0 & \textbf{98.6} \\
			\bottomrule
		\end{tabular}
		}
        \vspace{-0.3cm}
	\end{table}

	Tables \ref{VisDA-C}, \ref{Office-31}, and \ref{Digits} brief the classification performance of ElimPCL on three major datasets.
	ElimPCL outperforms current SOTA approaches in terms of average classification accuracy.
	Eliminating noise accumulation issue in the early training time helps ElimPCL perform well.
	Especially in settings with significant domain shifts, ElimPCL demonstrates strong performance improvements.
	For example, on \textbf{\textit{Office-31}} D$\rightarrow$A, ElimPCL exceeds SHOT\cite{liang2020we} by $+5.4\%$ and A$^{2}$Net\cite{xia2021adaptive} by $+3.4\%$.
	For \textbf{\textit{Office-31}} W$\rightarrow$A, although ElimPCL does not achieve the highest accuracy, it also shows a $+3.0\%$ gain over SHOT\cite{liang2020we}, ranking second only to UTR\cite{pei2023uncertainty}.
	On the challenging large-scale \textbf{\textit{VisDA-C}} dataset, ElimPCL achieves the highest per-class average accuracy, 
	and outperforms SOTA on the `\textit{car}' and `\textit{mcycl}' classes by $+2.2\%$ and $+2.3\%$, respectively. 
	This may be because ElimPCL effectively excludes high-noise samples before adaptation and enhances the separability of hard samples during adaptation.
	The \textbf{\textit{Digits}} dataset, being relatively simple with less severe domain shifts, also shows that ElimPCL achieves better classification accuracy compared to other SFDA methods.
	On the other hand, ElimPCL obtains competitive performance compared to standard UDA methods, even without direct access to the source domain data.
    
	\subsection{Ablation Study}
\begin{figure}[H]
		\begin{center}
			\includegraphics[width=0.48\textwidth, height=0.17\textwidth]{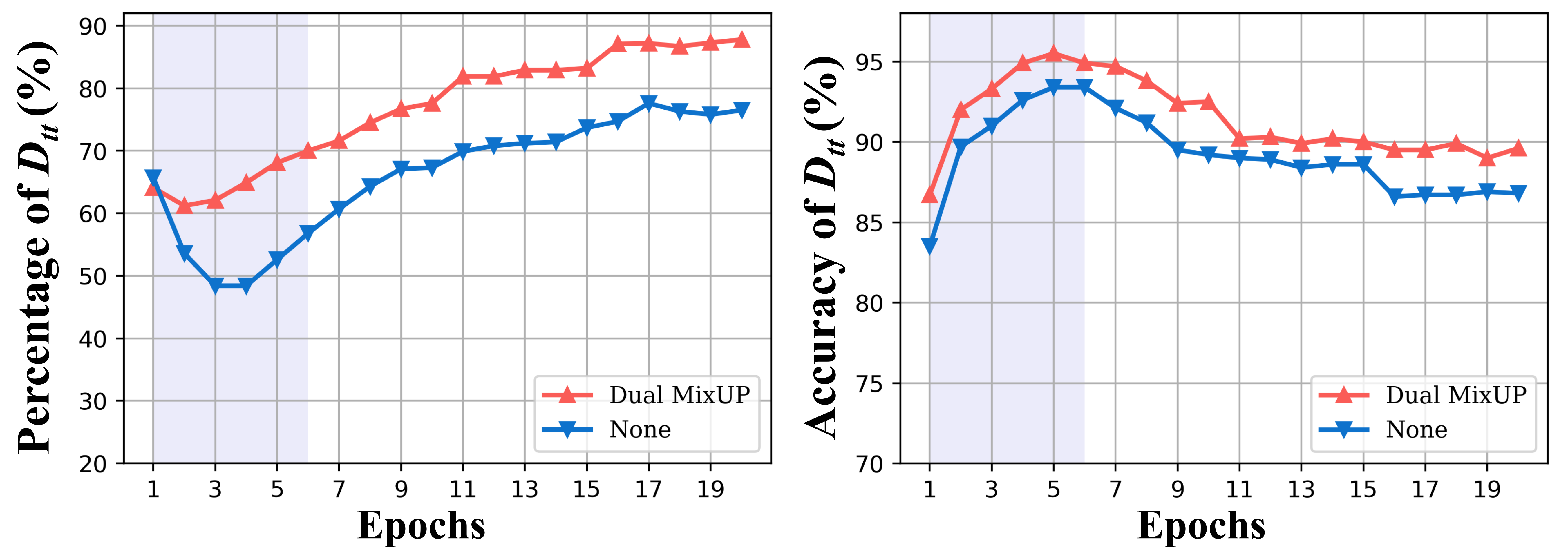}
		\end{center}
            \vspace{-0.3cm}
		\caption{Accuracy and percentage of the trustworthy subset $|D_{tt}|$ within the entire target domain $|D_{t}|$ on \textbf{\textit{VisDA-C}}.}
		\label{select_ratio}
        \vspace{-0.5cm}
	\end{figure}

	As shown in Table \ref{Ablation}, without prototype consistency filtering module, performance drops the most, as much as $-11.4\%$.
	It is obvious that prototype consistency filtering is the most significant tool for excluding high-noise samples from training, yet it is often overlooked by other approaches.
    We further analyze the contribution of Dual MixUP module. Fig.\ref{select_ratio} dem-
        \begin{table}[ht]
		\centering
		\caption{Ablation study results on \textbf{\textit{VisDA-C}}.}
		\label{Ablation}  
		\scriptsize
		\resizebox{0.48\textwidth}{0.048\textheight}{
		\begin{tabular}{ccc>{\columncolor{lighterpurple}}c}
			\toprule	
			\textbf{\makecell[c]{Prototype Consistency \\Filtering}} & \textbf{\makecell[c]{Dual \\MixUP}} & \textbf{\makecell[c]{Co-learning with \\ImageNet Networks}} & \makebox[0.051\textwidth][c]{\textbf{ACC.(\%)}}\\
			\noalign{\smallskip}\hline\noalign{\smallskip}
			\checked & \checked & \checked & \textbf{87.9$~~~~~~$} \\
			$\times$ & \checked & \checked & 76.5\textbf{$\text{\tiny\color{red}{\textbf{($\downarrow$11.4)}}}$}\\
			\checked & $\times$ & \checked & 82.8\textbf{$\text{\tiny\color{red}{($\downarrow$5.1)$~$}}$} \\
			\checked & \checked & $\times$ & 85.1\textbf{$\text{\tiny\color{red}{($\downarrow$2.8)$~$}}$} \\
			\bottomrule
		\end{tabular}
		}
        \vspace{-0.4cm}
	\end{table}
    
    \begin{figure}[H]
		\begin{center}
			\includegraphics[width=0.45\textwidth, height=0.2\textwidth]{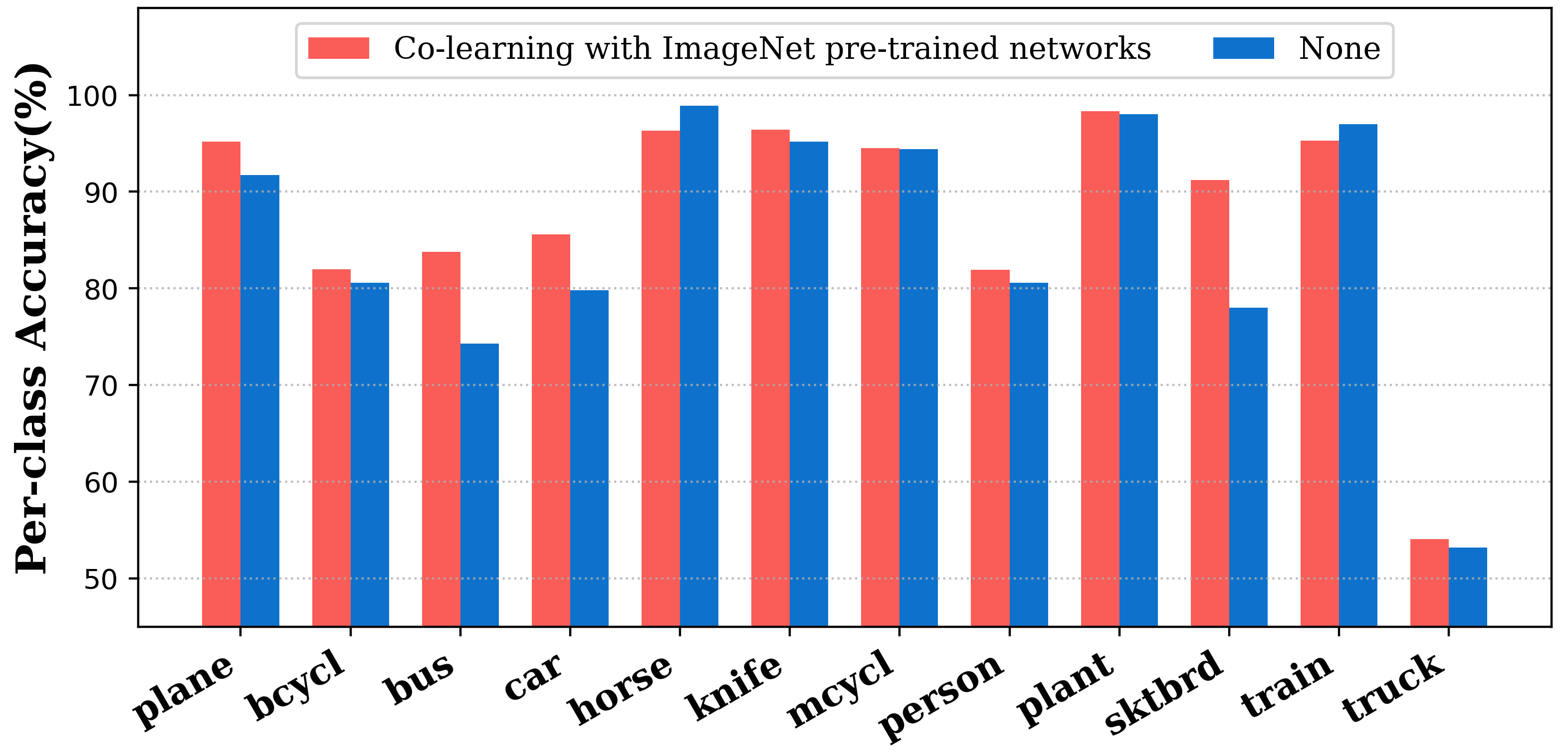}
		\end{center}
            \vspace{-0.3cm}
		\caption{Per-class accuracy on \textbf{\textit{VisDA-C}}.}
		\label{select_acc}
        \vspace{-0.3cm}
	\end{figure}
    
    \noindent onstrates that Dual MixUP technique facilitates more samples transitioning into the trustworthy subset, prompting the source model to generate higher-quality pseudo-labels. It is clear that Dual MixUP enhances the separability of hard samples in the feature space, thus alleviating the noise propagation effect.
	To validate the effectiveness of co-learning with ImageNet pre-trained network, Fig.\ref{select_acc} presents the per-class accuracy on the \textbf{\textit{VisDA-C}} dataset.
	This strategy boosts the accuracy of `\textit{bus}' by $+9.5\%$ and `\textit{sktbrd}' by $+13.2\%$, which are heavily affected by domain shifts with lower initial accuracy. This is because co-learning with ImageNet networks mitigates overfitting to the source domain and helps ElimPCL to extract more generalized features, especially for some hard samples.

	\subsection{t-SNE Visualization}
	\begin{figure}[H]
		\centering
		\subfloat[\textbf{\textit{Digits}} USPS$\rightarrow$MNIST]{\includegraphics[width=0.462\textwidth, height=0.198\textwidth]{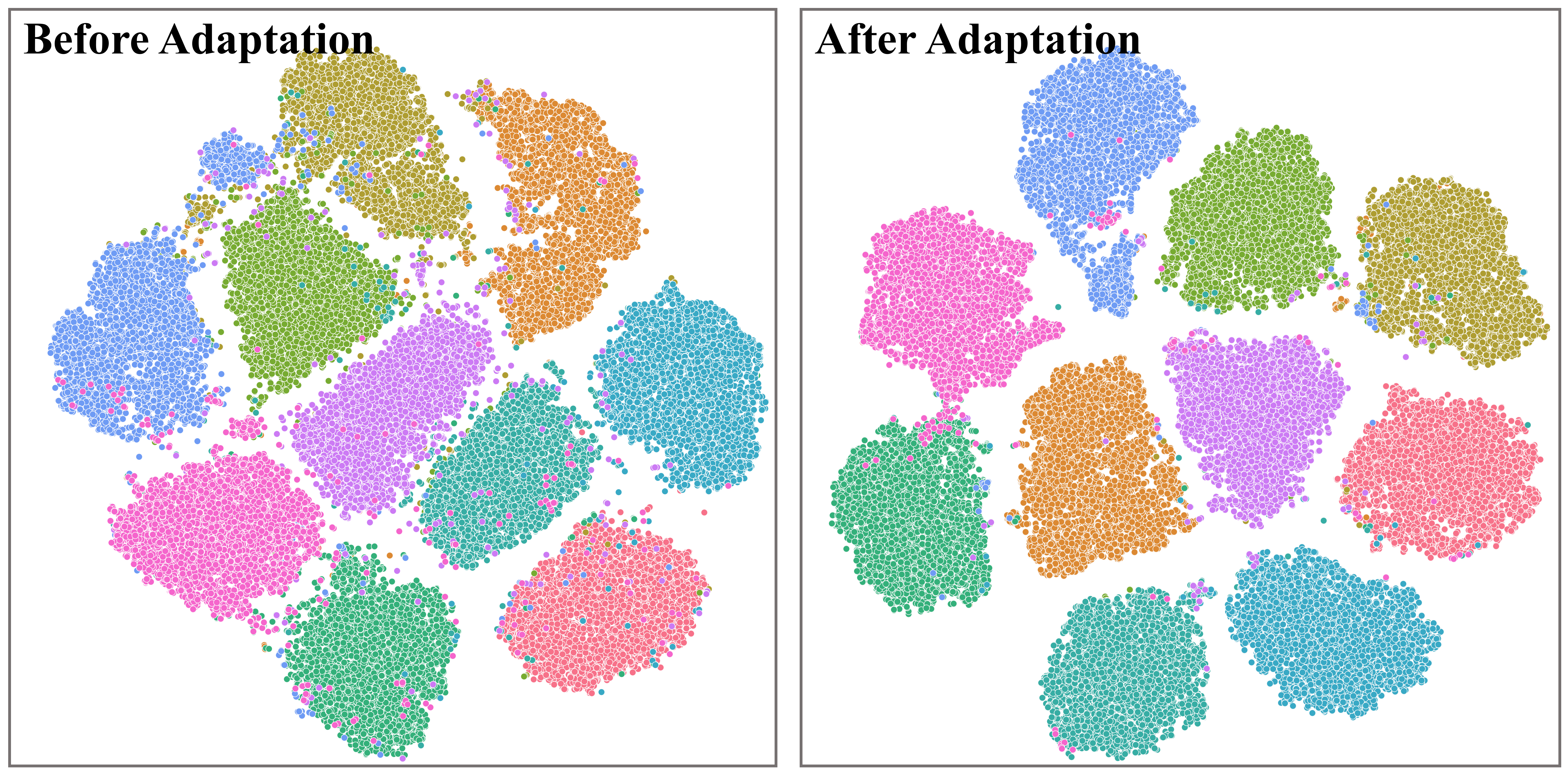}}

		\subfloat[\textbf{\textit{Office-Caltech}} Caltech$\rightarrow$Amazon]{\includegraphics[width=0.462\textwidth, height=0.198\textwidth]{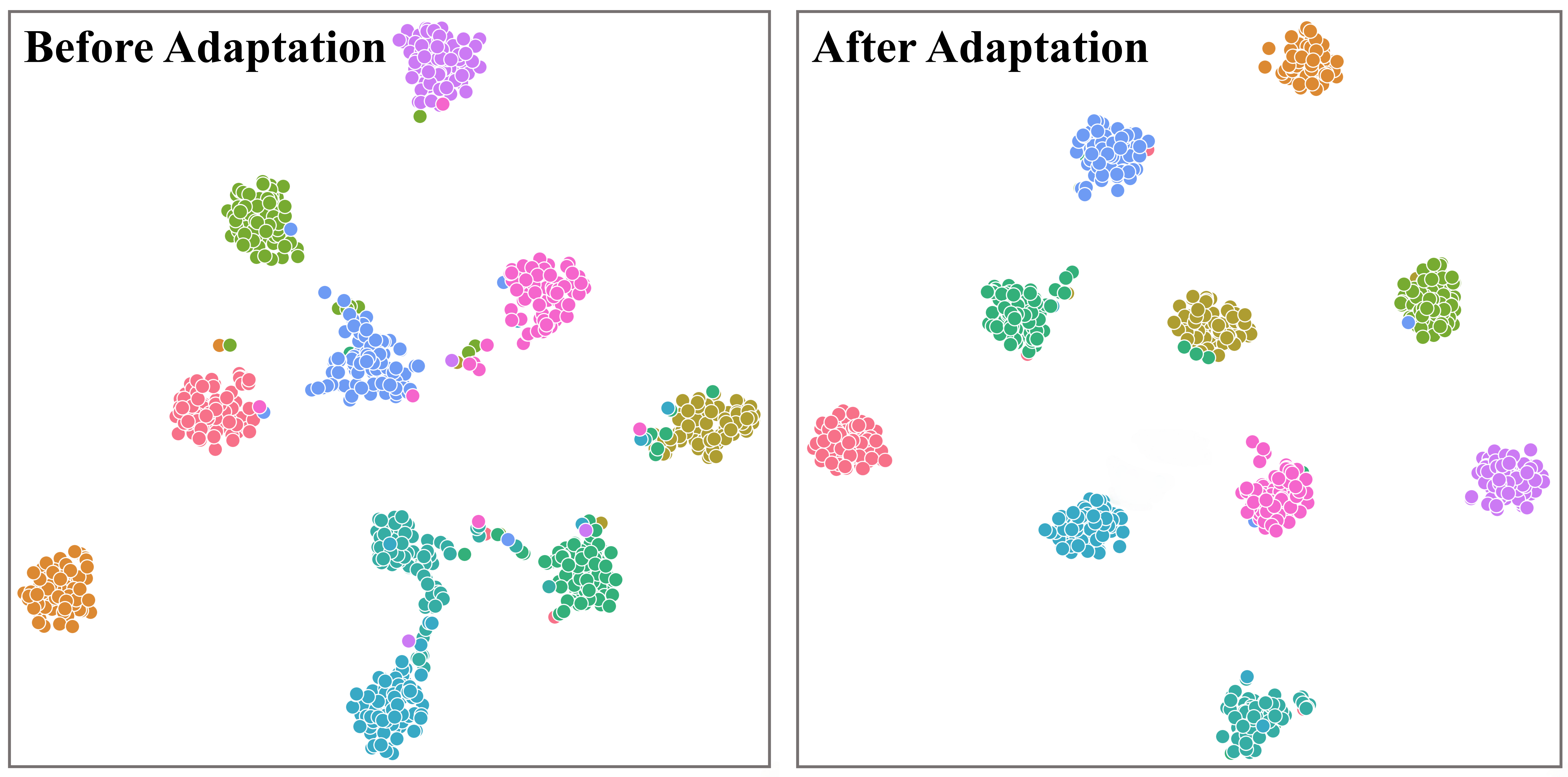}}
            \vspace{-0.1cm}
		\caption{Feature visualization with ElimPCL.}
		\label{t-SNE}
        \vspace{-0.3cm}
	\end{figure}

	As shown in Fig.\ref{t-SNE}, the feature distribution is highly disordered before domain adaptation.
	However, with the help of ElimPCL, samples with similar semantic features are grouped into tightly clustered regions, 
	and the classification boundaries between clusters become much more distinct.
	Taking \textbf{\textit{Digits}} dataset as an example, some of the blue samples are misaligned before domain adaptation, but eventually are corrected by ElimPCL.
	This indicates that ElimPCL perfectly solves the noise accumulation issue, avoiding mislabeled information from spreading to neighbor samples.

	\section{Conclusion}
	In this paper, we have proposed a novel ElimPCL method for SFDA.
	The ElimPCL focuses on solving how to iteratively refine and filter noisy pseudo-labels in the target domain,
	thus progressively fine-tuning the pre-trained source model using trustworthy subset samples.
	Extensive experiments confirm that ElimPCL effectively eliminates the noise accumulation issue by excluding hard samples from training before adaptation.
	In the future, we will be interested in exploring more lightweight label denoising techniques to reduce the overhead of label refinement process.
	We believe that ElimPCL could provide a new perspective to break through the performance bottleneck of current SOTA SFDA methods.

    \bibliographystyle{IEEEtran}
	\bibliography{PCL}
\end{document}